\def\year{2020}\relax
\documentclass[letterpaper]{article} 
\usepackage{booktabs}
\usepackage{blindtext}
\usepackage{enumitem}
\usepackage{aaai20}  
\usepackage{times}  
\usepackage{helvet} 
\usepackage{courier}  
\usepackage[hyphens]{url}  
\usepackage{graphicx} 
\usepackage{graphicx}  
\usepackage{amsmath}
\title{Fast Cross-domain Data Augmentation through Neural Sentence Editing}
\author{\Large Guillaume Raille\textsuperscript{\rm *}\textsuperscript{\rm 1}, Sandra Djambazovska\textsuperscript{\rm *}\textsuperscript{\rm 2}, Claudiu Musat\textsuperscript{\rm 3}.\\ 
\textsuperscript{\rm *} equal contribution\\
\textsuperscript{\rm 1} guillaume.raille@gmail.com\\ 
\textsuperscript{\rm 2} sandra.djambazovska@gmail.com\\
\textsuperscript{\rm 3} claudiu.musat@swisscom.com\\
}
\begin{document}

\maketitle

\begin{abstract}
Data augmentation promises to alleviate data scarcity. This is most important in cases where the initial data is in short supply. This is, for existing methods, also where augmenting is the most difficult, as learning the full data distribution is impossible. For natural language, sentence editing offers a solution - relying on small but meaningful changes to the original ones. Learning which changes are meaningful also requires large amounts of training data. We thus aim to learn this in a source domain where data is abundant and apply it in a different, target domain, where data is scarce - cross-domain augmentation.

We create the Edit-transformer, a Transformer-based sentence editor that is significantly faster than the state of the art and also works cross-domain.  We argue that, due to its structure, the Edit-transformer is better suited for cross-domain environments than its edit-based predecessors. We show this performance gap on the Yelp-Wikipedia domain pairs. 
Finally, we show that due to this cross-domain performance advantage, the Edit-transformer leads to meaningful performance gains in several downstream tasks.
\end{abstract}

\section{Introduction}
Models are a reflection of the data used for training them. In many cases, the available data is insufficient, leading to the creation of whole fields dedicated to low-data regimes \cite{zoph2016transfer,yang2017transfer}. Data Augmentation collectively designates methods aimed at generating additional, synthetic training data. 

The complexity of data augmentation ranges from simple techniques that duplicate certain data samples to balance a dataset \cite{10.1007/978-3-540-39804-2_12} to methods that learn the data distributions and sample novel points within them \cite{antoniou2017data}.
In the latter, the novel, synthetic data needs to be sufficiently close to the original samples to still be considered part of the same distribution. 

Downstream systems then rely on either the synthetic data alone or on combinations or real and synthetic data points to learn from.
When the original and synthetic data are used jointly, the synthetic data needs to be different enough from the original ones so that these variations lead to a better generalization capacity of the downstream methods. For instance, for NLP tasks \cite{bowman2015generating} the novel texts or phrases need to be novel realistic ones that can first pass as real human language and then seem as coming from the same domain as the originals.

Data augmentation is thus a form of transfer. 
At the very least, the synthetic sentences need to still be proper language but other criteria can be added - for instance keeping certain relations in the original sentence \cite{DBLP:journals/corr/HuYLSX17}. The novelties in the synthetic data are effectively enriching the original data and making them more useful for further processing.
A desirable property of data augmentation techniques is the ability to work in cross-domain settings. If, for instance training data is abundantly available in a source domain, the goal is to use this abundance to generate data samples in a resource-scarce target domain. While recent work addresses this for image generation \cite{yamaguchi2019effective,auggan}, for text the challenge is still unsolved.

Many flavors of data augmentation have been proposed. In particular methods based on Variational Autoencoders (VAE) carry the promise of adding a significant amount of variation with respect to the original samples, while staying close to the initial data distribution. The randomness of the perturbations performed on the embedded samples ensures the existence of novelty.
This family of methods are plagued by the difficulty of meeting all the expectations simultaneously. More often than not the results of the augmentation process are unsatisfactory, thus conditioning on secondary criteria, such as topics \cite{DBLP:journals/corr/abs-1903-07137} is needed.

An alternative is to focus the changes in a handful of edits. This idea is the starting point of the Neural Editor \cite{DBLP:journals/corr/abs-1709-08878}.
While the results of the Neural Editor have an easier time passing as human language, they come with two major downsides. Firstly, the method takes long to train and the time needed can, depending on the task, be measured in GPU days.
Moreover, in cross domain settings, the method performs very differently depending on the relation between the source and target domains.
We show in this work that the transition from the same-domain to cross-domain augmentation changes the quality of the output of the Neural Editor entirely, making it unusable for cross-domain settings.
The applicability of a data augmentation method in cross-domain scenarios is key for its usage in low-data regimes. If data is already plentiful in a domain, the need for augmentation decreases correspondingly.

Based on these observations, in this paper we introduce a method that tackles both aforementioned problems simultaneously. Our proposed method, the Edit-Transformer, replaces the recurrence within the Neural Editor with an attention-based component inspired by the Transformer architecture \cite{DBLP:journals/corr/VaswaniSPUJGKP17}.

The Edit-Transformer inspired by the Neural Editor consist in learning a consistent and controlled edition model over a large training dataset. Applying this model at inference time on an unrelated smaller dataset allows to increase its size and improve its performance on a given classification task. While learning editions based on a combination of sequence-to-sequence model and Variational Auto-Encoder was already introduced in the Neural Editor, several necessary improvements are implemented in the Edit-Transformer in order to apply a similar model in the context of data augmentation.

We show that the Edit-transformer has a great speed advantage over the Neural Editor baseline - with nearly three times faster training and five times faster inference. Moreover, we show empirically and then argue that, due to its structure, the Edit-Transformer is better suited for cross-domain data augmentation. 

This property makes it a great fit for real applications. Whereas most text augmentation techniques focus on how the synthetic sentences look, we analyze their impact on real-world applications. We thus go beyond a qualitative evaluation of the augmentation outcome and prove a real-world benefit of the method. This stems from the robustness of the method across domains.

We show that synthetic data from the Edit Transformer benefits three different binary classification datasets, including two datasets - Stanford Sentiment Treebank (SST-2) \cite{socher-etal-2013-parsing} and the Subjectivity dataset \cite{pang-lee-2004-sentimental} - that are common targets for data augmentation. The benefits are measured in terms of improvements on the performance of the downstream task, not directly using the characteristics of the synthetic data points.
Importantly, on Subj we find that the impact of the synthetic data even exceeds the impact of an equivalent quantity of additional real data.
We argue that the method is complementary to other state of the art data augmentation techniques, like back translation \cite{DBLP:journals/corr/FadaeeBM17} or EDA \cite{DBLP:journals/corr/abs-1901-11196}.

\section{Related Work}

Popular approaches to perform data augmentation on textual data consists in performing simple editions such as replacing adjectives, translating some part of sentences or randomly removing some words \cite{DBLP:journals/corr/abs-1901-11196}. Others rely on the use of "back-translation" (translation to a foreign language and then back to the original language) in order to generate new text samples \cite{DBLP:journals/corr/FadaeeBM17}. In this paper, we propose a complementary and compatible alternative to existing techniques and use generative models to create synthetic text samples that enlarge a given dataset.

Text generating models is a vast field that has been actively studied in the past. Two popular base approaches stands out when performing text generation: the Variational Auto-Encoder (VAE) with \cite{Bowman_2016} or \cite{Wang_2019} and Generative Adversarial Network(s) (GAN) with \cite{Iyyer_2018} or \cite{xu2018dpgan}. In the specific context of data augmentation, several additional requirements over regular text generation needs to be fulfilled in order to successfully use a text generating model. One of these requirements is the possibility to control the generation in some way to prevent wrong labeling of the augmented dataset. One way this has been achieved in the past while using the VAE was by adding a controlled latent variable in the latent space \cite{hu2017controlled}. Another attempt at generating text within a given range of changes has been performed by using a VAE to generate "editions" between two text samples instead of directly generating a new text sample \cite{DBLP:journals/corr/abs-1709-08878}. Restraining the range of edition of this approach to a rather small subset of possible editions reduce the risk of the generation process to affect the semantic meaning of the input text hence the labels are assumed to be preserved. 

Convergence issues of the VAE when generating text samples from random vectors and the lack of solid control over the generation convinced us to focus on performing data augmentation by generating "text editions" rather than directly text samples. For that reason our work is based on the neural-editor \cite{DBLP:journals/corr/abs-1709-08878}. However unlike previous approach, our model has been made viable to perform textual data augmentation by allowing cross-domain adaptability and increasing its performance and speed.

\section{Method}

Our data augmentation method consists in applying generated "text editions" that don't significantly change the meaning of a given text sample. Similarly to the neural-editor we focused our work at a sentence level. At a glance, performing data augmentation with our method consists in first generating one or several edition vectors for each input sentence and then constructing new sentences based on "edition vectors" and the corresponding input sentences.

Two components are key in allowing us to perform such editions: the first one is a sequence-to-sequence model to perform the encoding of an original sentence and decoding into new synthetic sentences. The second one is a Variational Auto-Encoder (VAE) to learn and encode non-breaking editions and randomly generates possible "edit vectors" for each sentences. The latent representation of the Variational Auto-Encoder comes to alter the decoder or generating part of the sequence-to-sequence model in order to correctly perform sentence editions.

The probabilistic model which properly describes sentence editing can be seen as follow. Given a pair of sentences from a given dataset $x$ and $x^\prime$ and a vector representation of an edition $z$ that transform the sentence $x^\prime$ into $x$, we want to maximise the probability of re-generating the original sentence $x$ given $x^\prime$ and $z$. Formally we want to maximise the following probability distribution:

\begin{equation}
	p_{edit}(x | x^\prime, z)
\end{equation}

where $p_{edit}$ corresponds to the probability distribution of generating a sentence corresponding to a "correct" modified sentence $x$ given an original sentence $x^\prime$ and a vector of edition $z$.

\subsection{Edit vector representation}
The edit vector representation $z$ is learned at training time using another probabilistic model based on the Variational Auto Encoder (VAE) \cite{Bowman_2016}. In this case we were maximizing the following probability distribution:

\begin{equation}
	q(z | x^\prime, x)
\end{equation}

where $q$ is the probability distribution of generating the latent representation $z$ conditioned on a pair of sentence $x$ and $x^\prime$ that will lead the overall model to output the correct sentence $x$ given $x^\prime$.


Similarly to the Variational Auto-Encoder the $KL$ divergence between the edit vector distribution and a prior distribution $p(z)$ is minimized through the overall objective function. As in the neural-editor, we used a prior distribution $p(z)$ that conveniently lead the KL divergence term to a constant with respect to the parameters of the model hence removing the KL divergence from our final objective.

In order to focus on non "meaning-breaking" editions, we limited the range of pair of sentences on which edition were learned to those constrained in within a jaccard distance of less than 0.5. We made a similar assumption as in the neural-editor that sentences with a close jaccard distance are likely to have a similar meaning. Furthermore limiting the range of editions also reduced consequently the size of the training dataset hence the training time.

\subsection{Training Phase}

During the training phase of our model, the objective is to have a generated sentence that is grammatically correct and that respect the desired editions specified through the "edit vectors". In order to do so, a pair of sentences $x$ and $x^\prime$ is sampled from the training dataset and an edition vector $z$ is sampled from the corresponding VAE distribution (using the reparametrization trick introduced in \cite{Bowman_2016}). The model will during training minimize the difference between the generated sentence and the expected sentence $x$ using the objective function described in Equation \ref{neural-editor-obj}:

\begin{equation} \label{neural-editor-obj}
	E_{z \sim q(z|x^\prime,x)}[logp_{edit}(x | x^\prime,z)] - KL(q(z|x^\prime,x) \Vert p(z))
\end{equation}

\subsection{Inference Phase}

At inference time, the model generates new sentences by first sampling an edit vector $z$ from the prior distribution $p(z)$ and then feeding it with a sentence to augment denoted $x^\prime$ to the decoder in order to generate an edited sentence $x$.

\subsection{Model Architecture}
Overall our model implementation is conceptually similar to the one used in the neural-editor. For this reason, we will primarily focus on the changes that we introduced over the previous architecture and the improvements they brought. An overview of our model architecture can be observed on figure \ref{fig:model_overiew}.

\begin{figure}[h!]
  \centering
    \includegraphics[width=0.385\textwidth]{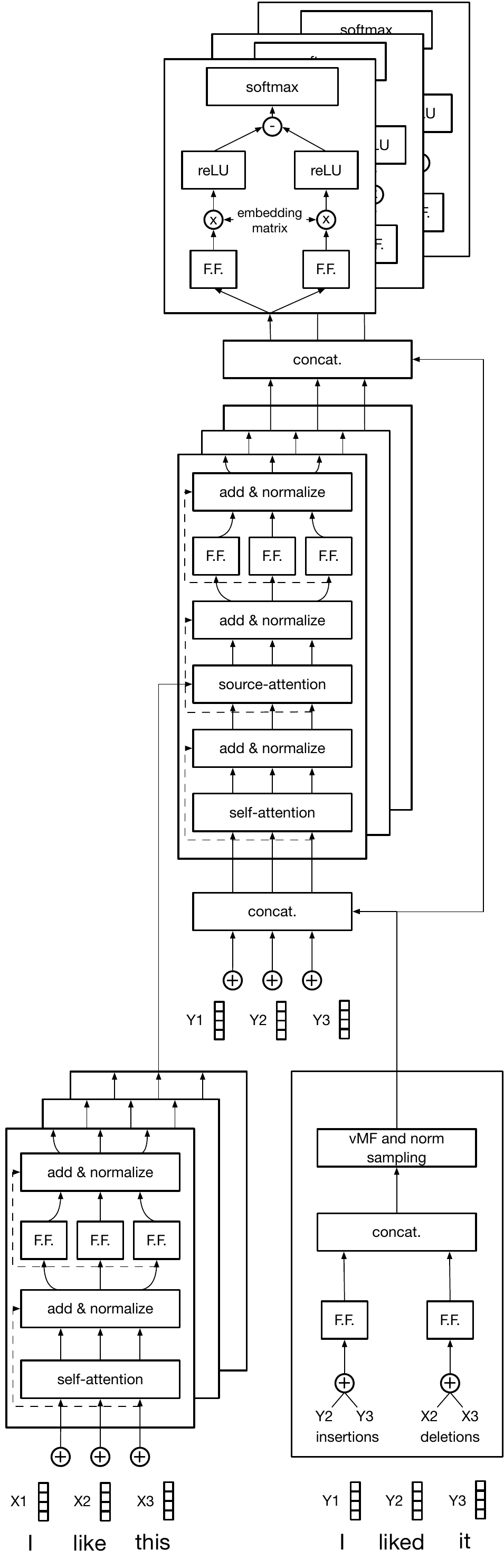}
  \caption{Model Architecture of the Edit Transformer}
  \label{fig:model_overiew}
\end{figure}

Figure \ref{fig:model_overiew} shows that the embedded vectors of the input sentence ($x_1$ to $x_3$) are encoded using a transformer encoder \cite{vaswani2017attention}. Whereas a latent representation based on the difference between an original sentence ($x_1$ to $x_3$) and a modified version of this sentence ($y_1$ to $y_3$) is used in a transformer decoder to learn to re-generate the original sentence ($x_1$ to $x_3$). This latent representation correspond to our edit vector $z$ based on a Variational Auto-Encoder model.

\subsubsection{Generalization}

In order to be applicable in a data augmentation context, our generative model once trained on a large corpus of text needs to perform inferences on a significantly smaller domain independent dataset. Previous implementations of "neural edition" were using trainable word embeddings to directly encode in the word embeddings information about editions. The inverse neural editor $q$ was designed in following way:

\begin{gather}
\label{q-detailed-eq}
		q(z_{dir} | x^\prime, x) = vMF(z_{dir};f_{dir},\kappa) \\
		q(z_{norm} | x^\prime, x) = Unif(z_{norm};[\tilde{f}_{norm}, \tilde{f}_{norm} + \epsilon])
\end{gather}

\bigskip

Where $f_{dir}$ and $f_{norm}$ are solely based on a sum of word embedding that differs between $x$ and $x^\prime$, $vMF$ and $Unif$ corresponds respectively to a "von Mises-Fisher" and a Uniform distribution and $\kappa$ and $\epsilon$ are fixed parameters. From equation \ref{q-detailed-eq}, the only trainable parameters that shapes the distribution $q$ from which is sampled the edition vector $z$ during training becomes the trainable word embedding encapsulated in $f$ in the previous equation.

This approach brings two major issues when inference are performed on an other domain. First it is impossible to add new word embedding at inference time since the word representation has changed in an unpredictable way. Hence if new words are encountered in a new dataset while performing inference, the model will never be able to re-generate them nor will they be well encoded by the encoder part of the model. Second, the embeddings won't be solely updated by the inverse neural editor $q$ during training as shown in the objective function but also by other parts of the model such as the sequence to sequence model. Experimental results shows that trainable embeddings tends to overfit the training dataset and leads to worst performance when encountering a slightly different domain at inference time.

To solve those issues and insure a better generalization of our implementation, we replaced trainable embeddings with pre-trained frozen embedding. Freezing the embedding not only reduced overfitting on the training dataset domain significantly, it also allowed us to add missing word embedding representation at inference time. Adding missing word embedding vectors at inference time significantly increased the performance of our implementation when new words appeared while augmenting a dataset. However, a consequence of using frozen embeddings is that no trainable parameters are left to train and adapt our inverse neural editor distribution $q$. In order to correctly learn a meaningful edit vector distribution we added a simple trainable linear layer that would translate the sum of frozen embedding to the vector $f$ used in equation \ref{q-detailed-eq}.

\subsubsection{Speed}

A major concern about previous implementation of the edition based model was the training time. In the context of data augmentation, many experiments and model variation were required before obtaining meaningful results. With the Neural Editor taking almost 2 GPU days to reach convergence, it would have been impossible to iterate and produce as many result as we did with a similar architecture. Slow training and inference performance also made the Neural-Editor too slow to make it a viable alternative to traditional data augmentation techniques. 

Previous implementation used a bi-LSTM encoder - decoder with attention approach to perform the text generation similar as in \cite{wu2016googles}. Most of the computational time and parameters of the model came from this sequence to sequence structure. In our implementation we decided to replace the RNN based approach with an attention based model similar to the one described in the transformer \cite{vaswani2017attention}.

Concretely no changes were introduced in the encoder part of the "transformer" which stayed a multilayer multi-headed attention encoder based on pre-trained GloVe word embeddings \cite{pennington-etal-2014-glove}. However in order to correctly account for the edit vector, we tuned the decoder. In the architecture of the neural-editor, editions were introduced in the LSTM decoder by concatenating at each time-step and at each layer the edit vector $z$ sampled from the inverse neural editor $q$ to each input vector of each LSTM cell. A similar approach would not work in the transformer due to the way "encoder-decoder attention" is computed:

\begin{equation}
	z_j = \sum_{i=0}^{d_s}(softmax(\frac{q_j \cdot K}{\sqrt{d_k}}) \cdot v_i)
\end{equation}

where $q$ a query vector is a linear projection coming from each token embeddings or the previous decoder layer in a sequence, $k$ of size $d_k$ and $v$ of size $d_v$ are the key and value vectors extracted as linear projections from the transformer encoder output. $z_j$ corresponds to the encoder-decoder attention representation of a token $j$ in a sequence of length $d_s$.

In particular, the values $v$ used in the "encoder-decoder attention" from which is computed the decoded representation of the sentence are coming from a linear projection of the encoder output values independent from the edit vector $z$. Since each output of each layer is mainly a weighted sum of the value $v$, most of the information coming from the edit vector $z$ is lost when concatenating it at each time-step input of the decoder.

The vanishing information problem appearing in the decoder was solved by concatenating the edit vector $z$, in the output of the decoder network. Two advantages comes out of this change: 

\begin{itemize}
	\item There is no loss of information in the different decoding layers of the architecture.
	\item The generator $p_g(x|z, out)$ can effectively take into consideration both untangled representations of the decoded past sequence $out$ and the edit vector $z$.
\end{itemize}

Different implementations of the "edit-transformer" decoder with different insertion of the edit vector $z$ were experimented. After trial and error, the insertion in the generator previously mentioned was kept as it was the one yielding the best performance on the validation set during training of the model.

The new architecture enabled significant improvements in speed and loss convergence while preserving the ability to retrieve and perform meaningful editions from a large dataset.

\section{Neural-editor vs Edit-transformer}

In this section, we will present the results we obtained when experimentally comparing the "neural-editor" with our new architecture the "edit-transformer" in terms of speed and performance over two different datasets.

\subsection{Datasets}
The train and test sets consist of pairs of sentences with a Jaccard distance lesser than 0.5. In order to efficiently match this pair of sentences, Latent Similarity Hashing (LSH) is performed on a corpus of individual sentences. All the named entities are replaced with tags using spacy\footnote{\url{https://spacy.io}}.
\begin{description}[font=$\bullet$~\normalfont\scshape]
\item \textbf{Yelp} Same as in \cite{DBLP:journals/corr/abs-1709-08878} we are using 4.5 million pairs of sentence for training, extracted from Yelp reviews\footnote{{Yelp dataset: }\url{ https://www.yelp.com/dataset_challenge}}. For testing we have 10'000 samples.
\item \textbf{Wikipedia} Each sentence in this dataset is from an articles with at least 20 page views from a dump of October 2013\footnote{Wikipedia dataset: https://blog.lateral.io/2015/06/the-unknown-perils-of-mining-wikipedia/}. We are using 13 million samples for training and 10'000 for testing. 

\end{description}
\subsection{Speed}

We first compared the training and inference times of the Neural Editor with those of the Edit Transformer. Both models were trained until convergence, with a patience of 50.000 steps, on the same Yelp train set. Then, when both models were trained, we computed the average inference time on one sample from the Yelp test set.

\subsubsection{Results}
In Table \ref{tab:speed} we present the results from the training and inference time of the Neural-editor and our model. Replacing the LSTM based model with a transformer resulted in a significant decrease of the training and inference time. The training time is reduced from 1 days and 15 hours to 10 hours. While the inference time is reduced by a factor of 5.

\begin{table}[h]
    \centering
    \begin{tabular}{c c c }
        \toprule
         &  Neural-editor & Edit-transformer\\
         \midrule
        train & 1d 15h & 10h\\
        inference & 31.3 ms  & 6.46 ms \\
        \bottomrule\\
    \end{tabular}
    \caption{Training and inference time NE vs ET}
    \label{tab:speed}
\end{table}{}

\subsection{Performance}
Moreover, we compared their performance on data from the same-domain and on out-of-domain data. We used the Yelp and Wikipedia corpus, for training and testing, and the bleu score and the loss as evaluation metrics. Both of the models were trained on the Yelp dataset. To compare them in the same domain, we used the test set from Yelp reviews. While for evaluating the models on out-of-domain data, we used the Wikipedia test set. We repeated the same process using the Wikipedia corpus. Both models were trained on Wikipedia and we tested their performance on the Yelp and Wikipedia test set.

\subsubsection{Results}
In Table \ref{tab:cross_domain} we show the results from the comparison. As expected, both models performed significantly better on same-domain data, compared to out-of-domain. However, the edit-transformer results are better than the neural-editor, both on same-domain and cross-domain data. Testing both models on same-domain data the edit-transformer is on average 3 times better than the neural-editor. While on out-of-domain it is at least 1.3 times better than the neural-editor. 

Due to insufficient performance of the neural-editor on out-of-domain data we continued our experiments only with the edit-transformer.

\begin{table*}[ht]
    \centering
    \begin{tabular}{ *{9}{c} }
        \toprule
        &  \multicolumn{4}{c}{Edit-transformer} & \multicolumn{4}{c}{Neural-editor }\\
        
         &  \multicolumn{2}{c}{Yelp test} & \multicolumn{2}{c}{Wikipedia test}& \multicolumn{2}{c}{Yelp test} & \multicolumn{2}{c}{Wikipedia test}\\
        \midrule
        train set & loss & bleu & loss & bleu  & loss & bleu & loss & bleu\\
        \midrule
        Yelp & 3.37 & 0.75 & 30.78 & 0.089 & 8.23 & 0.67 & 45.12 & 0.025\\
        
        Wikipedia & 33.46 & 0.062 & 4.75 & 0.7  & 43.79 & 0.032 & 20.23 & 0.104 \\
        \bottomrule
    \end{tabular}
    \caption{Performance of the Edit-transformer vs the Neural-editor}
    \label{tab:cross_domain}
\end{table*}{}

\section{Edit-transformer on downstream tasks}

\begin{figure}[h!]
  \centering
    \includegraphics[width=0.385\textwidth]{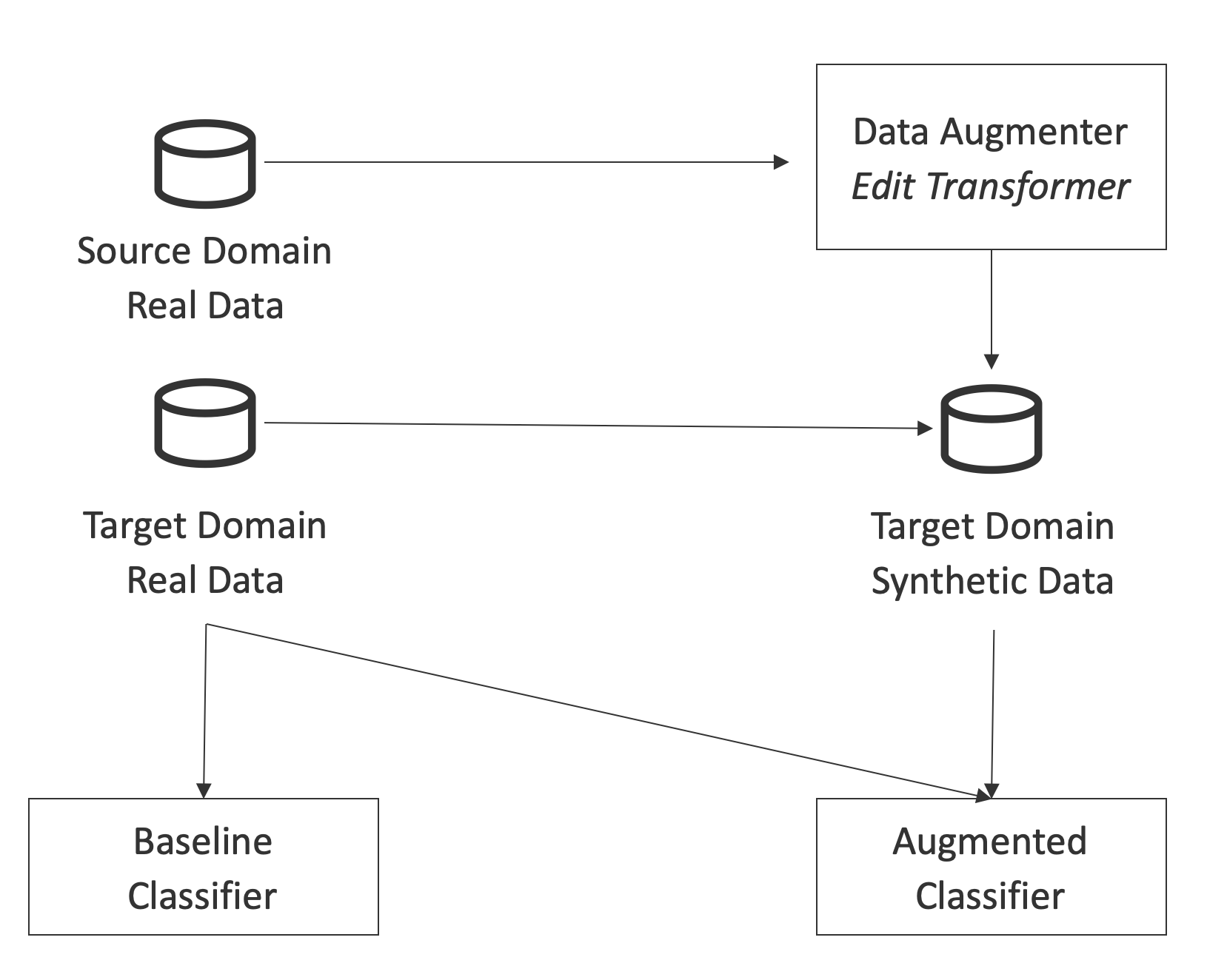}
  \caption{Cross-domain Data Augmentation Pipeline}
  \label{fig:model_overiew}
\end{figure}

In this section we will evaluate the performance of the edit-transformer on three different datasets. In all of the datasets described below, named entities were replaced with tags using spacy.

\subsection{Datasets}
\begin{description}[font=$\bullet$~\normalfont\scshape]
\item \textbf{Amazon reviews} is a binary classification task of amazon reviews \footnote{ Amazon dataset:

https://www.kaggle.com/bittlingmayer/amazonreviews} based on the following rule: 4 and 5 stars reviews were considered positive, 1 and 2 stars reviews were considered negative, 3 stars reviews were removed from the dataset. In the end we have 3.6 million training samples, 400'000 testing and 400'000 validation samples.
\item \textbf{SST-2} \cite{socher-etal-2013-parsing} is the Stanford Sentiment Treebank dataset with 10'000 samples labeled as positive or negative. For training we are using 8'000 samples, 1'000 for validation and 1'000 for testing.
\item \textbf{Subj} \cite{pang-lee-2004-sentimental} is a dataset where each sentence is classified as subjective or objective. It has 10'000 samples, 8'000 samples are used for training, 1'000 for testing and 1'000 for validation.
\end{description}
\subsection{Experimental setup}

In Figure \ref{fig:model_overiew} we have an overview of the cross-domain data augmentation pipeline. First we train the edit-transformer on the Yelp train set. Then using this model we generate new augmented data for each sample in the training set for SST-2, Amazon and Subj. We test the performance of the same model trained on the original and on the original plus augmented data, on an untouched test set.
We conduct the experiments using different fractions of the training set and we test the performance by augmenting each training sample once and twice. For the binary classification on Amazon reviews we are using an LSTM model. And for the SST-2 and Subj dataset, we are using a CNN model.

Although the label preservation was not modeled explicitly, implicitly the syntactic data will likely keep the label of the source. The results verify this assumption. 

All of the results below are the average accuracy over ten different seeds.
\subsection{Results on Subj task}
In Table \ref{tab:subj} we have the results on the Subj dataset. In the first and second column we have the percentage of the original data used for training, and the accuracy using this data, respectively. In the third and forth column we have the accuracy when augmenting each sample once and twice. 

We observe that we get constant improvement in all cases. We can also see that we get better performance by augmenting each sample twice, than by doubling the original data.

\begin{table}[h]
\centering
\begin{tabular}{c c c c }
\toprule
Training data used & Original data & N = 1 & N = 2 \\ 
\midrule
20\% & 88.616 & 90.074 & \bf 90.33\\
50\% & 90.327 & 91.38 & \bf 91.58\\
100\% & 91.443 & 92.66 & \bf 93.06\\
\bottomrule\\

\end{tabular}
\caption{Results on the Subj dataset}
\label{tab:subj}

\end{table}

\subsection{Results on SST-2}

In Table \ref{tab:sst} we show results on the SST-2 dataset. Same as before we performed the experiment using 20, 50 and 100 percent of the data. Here we notice that we get an improvement only when using 20 and 50$\%$ of the data. The reason for this could be because using 100\% of the original data is enough for the model to be able to generalize properly and by augmenting the data we are just adding samples that are very close to the real data so it does not improve the performance. 

\begin{table}[h]
\centering
\begin{tabular}{c c c c}
\toprule
Training data used & Original data & N = 1 & N = 2 \\ 
\midrule
20\% & 79.4308 & 80.223 & \bf 80.301\\
50\% & 81.116 & 81.5513 & \bf 81.5625\\
100\% & \bf 83.1584 & 82.2544 & 82.3214\\
\bottomrule\\
\end{tabular}
\caption{Accuracy on the SST-2 dataset}
\label{tab:sst}
\end{table}

\subsection{Results on Amazon reviews}
Here we were using 1,2,3 and 4 percent of the data, because the dataset is significantly larger than the previous two datasets. In Table \ref{tab:amazon} we have the results on the Amazon dataset. We show the accuracy and the standard deviation because the differences are smaller. We observe constant improvement in all cases.

\begin{table}[h]
\centering
\begin{tabular}{c c c c}
\toprule
Data  & Original data & N = 1 & N = 2 \\ 
\midrule
1\% & 85.75 $\pm$ 0.445 & 86.71 $\pm$ 0.282 & \bf 86.87 $\pm$ 0.240\\
2\% & 87.59 $\pm$ 0.245 & 88.25 $\pm$ 0.178 & \bf 88.46 $\pm$ 0.146\\
3\% & 88.62 $\pm$ 0.215 & 89.08 $\pm$ 0.302 & \bf 89.12 $\pm$ 0.145\\
4\% & 89.25 $\pm$ 0.180 & 89.66 $\pm$ 0.176 & \bf 89.66 $\pm$ 0.097\\
\bottomrule\\
\end{tabular}
\caption{Accuracy $\pm$ standard deviation on Amazon reviews dataset}
\label{tab:amazon}

\end{table}

\section{Conclusion}
In this paper we introduced the Edit Transformer - a next generation neural architecture to generate novel sentences through relevant edits. The most important change with respect to the prior work is the Edit Transformer's ability to function cross-domain - applying edits learned on a source domain with plentiful data to a data-constrained target domain. 
In addition, we show it is faster to both train and use than the Neural Editor that precedes it.

The results on several well-known data augmentation tasks show that the synthetic data the Edit Transformer generates helps in downstream tasks. In the case of the Subj dataset, the effect of the synthetic data on training a subjectivity classifier surpassed even that of an equivalent quantity of real data. 
These results make us confident that the Edit transformer is an important element in the text Data Augmentation toolbox.

\bibliographystyle{named}
\bibliography{references}

\end{document}